\documentclass[11pt,a4paper]{article}
\usepackage[hyperref]{emnlp-ijcnlp-2019}
\usepackage{times}
\usepackage{latexsym}
\usepackage{graphicx}
\usepackage{amsfonts}

\usepackage{url}

\aclfinalcopy 

\usepackage{subcaption}
\usepackage{amsmath}
\usepackage{pgfplots}
\usepackage{tikz}
\usepackage{multirow}

\usepackage{arydshln}

\usepackage{algorithm}
\usepackage[noend]{algpseudocode}
\usepackage{mathtools}
\usepackage{amssymb}

\include{plots}
\definecolor{bblue}{HTML}{4F81BD}
\definecolor{rred}{HTML}{C0504D}
\definecolor{ggreen}{HTML}{9BBB59}
\definecolor{ppurple}{HTML}{9F4C7C}
\definecolor{Dark scarlet}{HTML}{560319}
\definecolor{Forest green}{HTML}{1E4D2B}

\title{
Neural Architectures for Fine-Grained Propaganda Detection in News
}

\newcommand*{\affaddr}[1]{#1} 
\newcommand*{\affmark}[1][*]{\textsuperscript{#1}}

\author{Pankaj Gupta\affmark[1,2], Khushbu Saxena\affmark[1], Usama Yaseen\affmark[1,2], Thomas Runkler\affmark[1], Hinrich Sch\"{u}tze\affmark[2]\\ 
 \affaddr{\affmark[1]Corporate Technology, Machine-Intelligence (MIC-DE), Siemens AG  Munich, Germany}\\
  \affaddr{\affmark[2]CIS, University of Munich (LMU) Munich, Germany} \\
  {\tt pankaj.gupta@siemens.com  | pankaj.gupta@campus.lmu.de}\\
}

\date{}

\begin{document}
\maketitle
\begin{abstract}

This paper describes our system (MIC-CIS) details and results of participation in the fine-grained propaganda detection shared task 2019.   
To address the tasks of sentence (SLC) and fragment level (FLC) propaganda detection, we explore different neural architectures (e.g., CNN, LSTM-CRF and BERT) and 
extract linguistic (e.g., part-of-speech, named entity, readability, sentiment, emotion, etc.), layout and topical features. Specifically, we have designed 
multi-granularity and multi-tasking neural architectures to jointly perform both the sentence and fragment level propaganda detection. 
Additionally, we investigate different ensemble schemes such as majority-voting, relax-voting, etc. to boost overall system performance.   
Compared to the other participating systems, our submissions are ranked $3^{rd}$ and $4^{th}$ in FLC and SLC tasks, respectively. 

\end{abstract}

\section{Introduction}

In the age of information dissemination without quality control, it has enabled malicious users to spread misinformation via social media and aim individual users with propaganda campaigns to achieve political and financial gains as well as advance a specific agenda.     
Often disinformation is complied in the two major forms: fake news and propaganda, where they differ in the sense that the propaganda is possibly built upon true information (e.g., biased, loaded language, repetition, etc.).  

Prior works \cite{DBLP:conf/emnlp/RashkinCJVC17, DBLP:conf/emnlp/HabernalHPKPG17, DBLP:conf/aaai/Barron-CedenoMJ19} in detecting propaganda have focused primarily at document level, 
typically labeling all articles from a propagandistic news outlet as propaganda and thus, often non-propagandistic articles from the outlet are mislabeled.   
To this end, \newcite{EMNLP19DaSanMartino} focuses on analyzing the use of propaganda and detecting specific propagandistic techniques in news articles at sentence and fragment level, respectively and thus, promotes explainable AI. 
For instance, the following text is a propaganda of type `slogan'. 

\ \ \ \ \  {\small \texttt{Trump tweeted:} $\underbrace{\text``{\texttt{BUILD THE WALL!}"}}_{\text{slogan}}$}

{\bf Shared Task}: This work addresses the two tasks in propaganda detection \cite{EMNLP19DaSanMartino} of different granularities: 
(1) {\it Sentence-level Classification} (SLC), a binary classification that predicts whether a sentence contains at least one propaganda technique, and 
(2) {\it Fragment-level Classification} (FLC), a token-level (multi-label) classification that identifies both the spans and the type of propaganda technique(s).

{\bf Contributions}: 
{\bf (1)} To address SLC, we design an ensemble of different classifiers based on Logistic Regression, CNN and BERT, and leverage transfer learning benefits using the pre-trained embeddings/models from FastText and BERT. 
We also employed different features such as  linguistic (sentiment, readability, emotion, part-of-speech and named entity tags, etc.), layout, topics, etc. 
{\bf (2)} To address FLC, we design a multi-task neural sequence tagger based on LSTM-CRF and linguistic features to jointly detect propagandistic fragments and its type. 
Moreover, we investigate performing FLC and SLC jointly in a multi-granularity network based on LSTM-CRF and BERT. 
{\bf (3)} Our system (MIC-CIS) is ranked $3^{rd}$ (out of 12 participants)  and $4^{th}$ (out of 25 participants) in FLC and SLC tasks, respectively.

\begin{table*}[t]
\center
\renewcommand*{\arraystretch}{1.25}
\resizebox{.999\textwidth}{!}{
\setlength\tabcolsep{3.pt}
\begin{tabular}{r|c|l}

\multicolumn{1}{c|}{\bf Category}       &	\multicolumn{1}{c|}{\bf Feature}	& \multicolumn{1}{c}{\bf Description} \\ \hline

\multirow{8}{*}{\it Linguistic}       &		POS				&       part-of-speech tags using NLTk toolkit  \\ \cline{2-3}
						&		\multirow{2}{*}{NER} &       named-entity tags using spacy toolkit, selected tags:   \\
						&						&	  \{PERSON, NORP, FAC, ORG, GPE, LOC, EVENT, WORK\_OF\_ART, LAW, LANGUAGE\} \\ \cline{2-3}
						&		\multirow{2}{*}{character analysis}	&       count of question and exclamation marks in sentence\\ 
						&						&	capital features for each word: first-char-capital, all-char-capital, etc. \\ \cline{2-3}
						&		readability			&       readability and complexity scores using measures from textstat API\\  \cline{2-3}
						&		multi-meaning		&        sum of meanings of a word (grouped by POS) or its synonym nest in the sentence using WordNet\\ \cline{2-3}
						&		\multirow{2}{*}{sentiment} &       polarity (positive, negative, neural, compound) scores using spacy;  subjectivity using TextBlob;\\ 
						&							&	 max\_pos: maximum of positive,  max\_neg: max of negative scores of each word in the sentence \\ \cline{2-3}
						&		emotional			&       Emotion features (sadness, joy, fear, disgust, and anger) using IBM Watson NLU API \\   \cline{2-3} 
						&		loaded words		&       list of specific words and phrases with strong emotional implications (positive or negative)  \\   \hline 

\multirow{2}{*}{\it Layout}       	&	\multirow{2}{*}{sentence position}		&       categorized as [FIRST, TOP, MIDDLE, BOTTOM, LAST], where, FIRST: $1^{st}$,  \\
						&						&	TOP: $<$ 30\%, Middle: between 30-70\%, BOTTOM: $>$ 70\%, LAST: last sentence of document\\   \cline{2-3}
						&	\multirow{1}{*}{sentence length ($l$)}	&      categorized as [$=2, 2<l \le 4, 4<l \le 8, 8<l \le 20, 20<l \le 40, 40<l \le 60, l>60$] \\   \hline

{\it Topical}       			&	\multirow{2}{*}{topics} 	&       document-topic proportion using LDA, features derived using dominant topic (DT): [DT of current \\
						&						&	sentence == DT of document, DT of current sentence == DT of the next and previous sentences] \\  \hline

\multirow{2}{*}{\it Representation}       &	word vector			&      pre-trained word vectors from FastText ({\it FastTextWordEmb}) and BERT ({\it BERTWordEmb})\\ 
						&	sentence vector		&       summing word vectors of the sentence to obtain {\it FastTextSentEmb} and  {\it BERTSentEmb} \\  \hline

{\it Decision}       			&	relax-boundary		&    (binary classification) Relax decision boundary and tag propaganda if prediction probability $\ge \tau$\\  \hline

\multirow{2}{*}{\it Ensemble}       &		majority-voting	&    Propaganda if majority says propaganda. In conflict, take prediction of the model with highest F1\\  \cline{2-3}
     						 &		relax-voting	&    Propaganda if $\mathcal{M} \in [20\%, 30\%, 40\%]$ of models in the ensemble says propaganda.  

\end{tabular}}
\caption{Features used in SLC and FLC tasks}
\label{tab:features}
\end{table*}

\begin{figure*}[t]
\centering
    \includegraphics[scale=0.625]{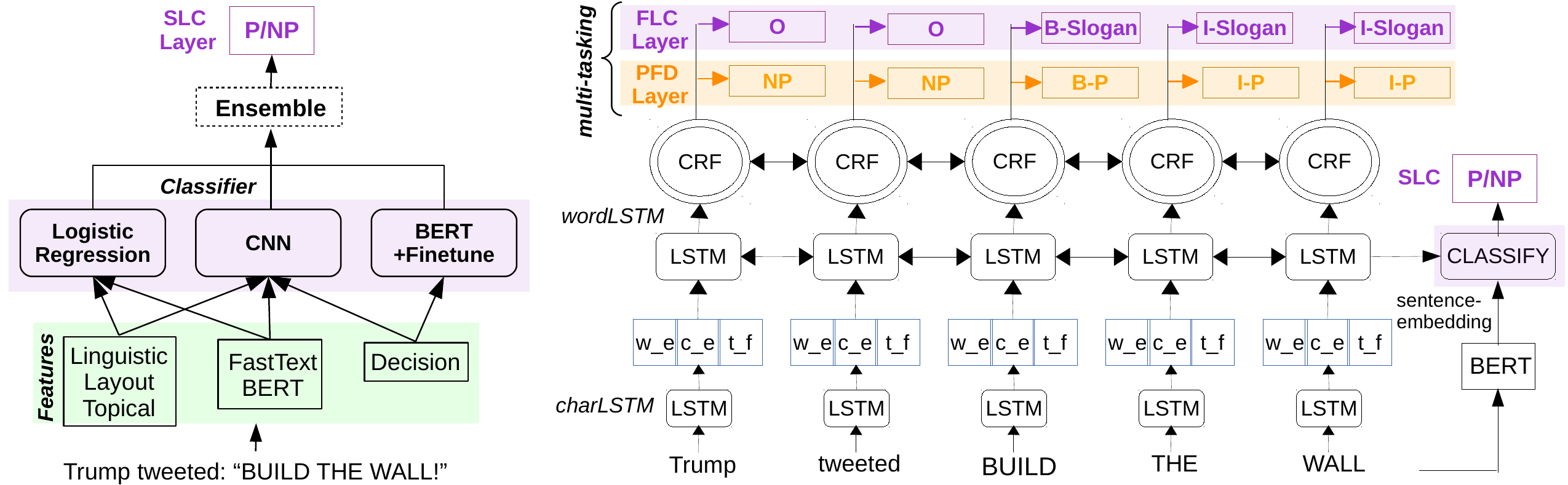}
    \caption{(Left): System description for SLC, including features, transfer learning using pre-trained word embeddings 
      from FastText and BERT and classifiers: LogisticRegression, CNN and BERT fine-tuning. 
      (Right): System description for FLC, including multi-tasking LSTM-CRF architecture consisting of Propaganda Fragment Detection (PFD) and FLC layers. 
Observe, a binary classification component at the last hidden layer in the recurrent architecture that jointly performs PFD, FLC and SLC tasks (i.e., multi-grained propaganda detection). 
Here, P: Propaganda, NP: Non-propaganda, B/I/O: Begin, Intermediate and Other tags of BIO tagging scheme.}
    \label{fig:SLCFLCArchitecture}
\end{figure*}

\section{System Description}

\subsection{Linguistic, Layout and Topical Features}

Some of the propaganda techniques \cite{EMNLP19DaSanMartino} involve word and phrases that express strong emotional implications, exaggeration, minimization, doubt, national feeling, labeling , stereotyping, etc. 
This inspires\footnote{some features from \url{datasciencesociety.net/detecting-propaganda-on-sentence-level/}} us in extracting different features (Table \ref{tab:features}) including the complexity of 
text, sentiment, emotion, lexical (POS, NER, etc.), layout, etc. To further investigate, we use topical features (e.g., document-topic proportion) \cite{DBLP:journals/jmlr/BleiNJ03, pankajgupta:2019iDocNADEe, pankajgupta:2018RNNRSM} at sentence and document levels in order to 
determine irrelevant themes, if introduced to the issue being discussed (e.g., {\it Red Herring}).  

For word and sentence representations, we use pre-trained vectors from FastText \cite{DBLP:journals/tacl/BojanowskiGJM17} and BERT \cite{DBLP:conf/naacl/DevlinCLT19}.

\subsection{Sentence-level Propaganda Detection}

Figure \ref{fig:SLCFLCArchitecture} (left) describes the three components of our system for SLC task: features, classifiers and ensemble. 
The arrows from features-to-classifier indicate that we investigate linguistic, layout and topical features in the two binary classifiers: LogisticRegression and CNN. 
For CNN, we follow the architecture of \newcite{DBLP:conf/emnlp/Kim14} for sentence-level classification, initializing the word vectors by FastText or BERT. 
We concatenate features in the last hidden layer before classification.   

One of our strong classifiers includes BERT that has achieved state-of-the-art performance on multiple NLP benchmarks. Following \newcite{DBLP:conf/naacl/DevlinCLT19}, 
we fine-tune BERT for binary classification, initializing with a pre-trained model (i.e., {\it BERT-base, Cased}). 
Additionally, we apply a decision function such that a sentence is tagged as propaganda if prediction probability of the classifier is greater than a threshold ($\tau$). 
We relax the binary decision boundary to boost recall, similar to \newcite{pankajgupta:CrossRE2019}.

{\bf Ensemble of Logistic Regression, CNN and BERT}: In the final component, we collect predictions (i.e., propaganda label) 
for each sentence from the three ($\mathcal{M}=3$) classifiers and thus, obtain $\mathcal{M}$ number of predictions for each sentence.  
We explore two ensemble strategies (Table \ref{tab:features}): majority-voting and relax-voting to boost precision and recall, respectively.

\subsection{Fragment-level Propaganda Detection}\label{sec:FLC}

Figure \ref{fig:SLCFLCArchitecture} (right) describes our system for FLC task, where we design sequence taggers \cite{ThangVu:2016CNNRNNRE, pankajgupta:2016TFMTRNN} in three modes: 
(1) {\it LSTM-CRF} \cite{DBLP:conf/naacl/LampleBSKD16} with word embeddings ($w\_e$) and character embeddings $c\_e$, token-level features ($t\_f$) such as polarity, POS, NER, etc.
(2) {\it LSTM-CRF+Multi-grain} that jointly performs FLC and SLC with FastTextWordEmb and BERTSentEmb, respectively. Here, we add binary sentence classification loss to sequence tagging weighted by a factor of $\alpha$.  
(3) {\it LSTM-CRF+Multi-task} that performs propagandistic span/fragment detection (PFD)  and FLC (fragment detection + 19-way classification).

{\bf Ensemble of Multi-grain, Multi-task LSTM-CRF with BERT}: 
Here, we build an ensemble by considering propagandistic fragments (and its type) from each of the sequence taggers. 
In doing so, we first perform majority voting at the fragment level for the fragment where their spans exactly overlap. In case of non-overlapping fragments, we consider all.
However, when the spans overlap (though with the same label),  we consider the fragment with the largest span.

\section{Experiments and Evaluation}

{\bf Data}:  While the SLC task is binary, the FLC consists of 18 propaganda techniques \cite{EMNLP19DaSanMartino}.
We split (80-20\%) the annotated corpus into 5-folds and 3-folds for SLC and FLC tasks, respectively. 
The development set of each the folds is represented by dev (internal); however, the un-annotated corpus used in leaderboard comparisons by dev (external).  
We remove empty and single token sentences after tokenization.

{\bf Experimental Setup}: 
We use PyTorch framework for the pre-trained BERT model ({\it Bert-base-cased}\footnote{\url{github.com/ThilinaRajapakse/pytorch-transformers-classification}}), 
 fine-tuned for SLC task. 
In the multi-granularity loss, we set $\alpha = 0.1$ for sentence classification based on dev (internal, fold1) scores. We use BIO tagging scheme of NER in FLC task. 
For CNN, we follow \newcite{DBLP:conf/emnlp/Kim14} with filter-sizes of [2, 3, 4, 5, 6], 128 filters and 16 batch-size. 
We compute binary-F1and macro-F1\footnote{evaluation measure with strict boundary detection} 
\cite{DBLP:journals/bmcbi/TsaiWCLHHSH06} in SLC and FLC, respectively on dev (internal).   

\begin{table}[t]
\center
\small
\renewcommand*{\arraystretch}{1.25}
\resizebox{.497\textwidth}{!}{
\setlength\tabcolsep{3.pt}
\begin{tabular}{r|c|r|c}
 \multicolumn{2}{c|}{{\bf Task: SLC} (25 participants)}					&   \multicolumn{2}{c}{{\bf Task: FLC} (12 participants)} \\ \hline
\multicolumn{1}{c|}{\bf Team}  		&    $F1$ \ / \    $P$ \  / \   $R$     			        &  \multicolumn{1}{c|}{\bf Team}      &     $F1$ \ / \    $P$ \  / \   $R$   \\ \hline
{\it ltuorp}						& {\bf .6323} / .6028	 / .6649			&   {\it newspeak}       &       {\bf .2488} / .2863 /  {\bf .2201} \\
{\it ProperGander}					& .6256  /	.5649   /  {\bf .7009}		&   {\it Antiganda}       &       .2267 / 	{\bf .2882} / .1869 \\
{\it YMJA}							&  .6249   / {\bf  .6253}   /  .6246   	&   {\bf MIC-CIS}       &         .1999   /	.2234   /  .1808 \\
{\bf MIC-CIS}				       &     .6231	/       .5736      /    .6819     	&   {\it Stalin}       &             .1453	/  .1921	/  .1169 \\
{\it TeamOne}					       &  .6183	/  .5779	/  .6649		&   {\it TeamOne}       &       .1311 / 	.3235 / 	.0822 
\end{tabular}}
\caption{Comparison of our system (MIC-CIS) with top-5 participants: Scores on Test set for SLC and FLC}
\label{tab:SLCFLCTestScores}
\end{table}

\begin{table}[t]
\center
\renewcommand*{\arraystretch}{1.25}
\resizebox{.495\textwidth}{!}{
\setlength\tabcolsep{3.pt}
\begin{tabular}{r|l|c|c}
\hline
& \multicolumn{2}{c|}{\bf Dev (internal), Fold1} &   \multicolumn{1}{c}{\bf Dev (external)} \\ 
& \multicolumn{1}{c|}{\bf Features}   &     $F1$ \ / \    $P$ \  / \   $R$   &     $F1$  \  /  \    $P$ \ /   \   $R$ \\ \hline
r1 &   {\it logisticReg} + TF-IDF     &   .569 / .542 / .598 &   .506 / .529 / .486  \\ \hline
r2  &  {\it logisticReg} + FastTextSentEmb    &   .606 / .544 / .685 &   .614 / .595 / .635  \\
    &  \   + Linguistic            &   .605 / .553 / .667 &   .613 / .593 / .633  \\
     & \   + Layout            &   .600 / .550 / .661 &   .611 / .591 / .633  \\
      &  \     + Topical            &   .603 / .552 / .664 &   .612 / .592 / .633  \\ \hline

r3   &  {\it logisticReg} + BERTSentEmb    &   .614 / .560 / .679 &   .636 / .638 / .635  \\
r4    &  \   + Linguistic, Layout, Topical            &   .611 / .564 / .666 &   .643 / .641 / .644  \\  \hline

r5   &   {\it CNN} + FastTextWordEmb            &   .616 / .685 / .559 &   .563 / .655 / .494  \\
r6    &  \   + BERTSentEmb               &   .612 / {\bf .693} / .548 &   .568 / .673 / .491  \\
r7    &  \   + Linguistic, Layout, Topical            &   .648 / .630 / .668 &   .632 / .644 / .621  \\ \hline

r8    &   {\it CNN} + BERTWordEmb            &   .610 / .688 / .549 &   .544 / .667 / .459  \\
r9     &  \   + Linguistic, Layout, Topical            &   .616 / .671 / .570 &   .555 / .662 / .478  \\ \hline

r10   &   {\it BERT} + Fine-tune ($\tau \ge .50$)            &   .662 / .635 / .692 &   .639 / .653 / .625  \\  

r11   &   {\it BERT} + Fine-tune ($\tau \ge  .40$)            &   .664 / .625 / .708 &   .649 / .651 / .647  \\  

r12   &    {\it BERT} + Fine-tune ($\tau \ge .35$)            &   .662 / .615 / .715 &   .650 / .647 / .654 \\ \hline \hline


\multicolumn{4}{c}{Ensemble of (r3, r6, r12) within Fold1}\\
r15   &     majority-voting  $|\mathcal M| > 50\%$           &   .666 / .663 / .671 &   .638 / .674 / .605  \\
r16  &   relax-voting, $|\mathcal M| \ge 30\%$             &   .645 / .528 / .826 &   .676 / .592 / .788  \\  \hline 

\multicolumn{4}{c}{{\bf Ensemble+} of  (r3, r6, r12) from each {\bf Fold1-5}, i.e., $|\mathcal M| = 15$}  \\
r17   &    majority-voting  $|\mathcal M| > 50\%$           &  		 	&   .666 / {\bf .683} / .649  \\
r18   &     relax-voting, $|\mathcal M| \ge 40\%$          &               	&   .670 / .646 / .696  \\
r19   &    relax-voting, $|\mathcal M| \ge 30\%$          &               	&   \underline{.673} / .619 / \underline{.737}  \\
r20   & \   + postprocess ($w$=10, $\lambda \ge .99$) &               	&   .669 / .612 / .737  \\
r21   & \   + postprocess ($w$=10, $\lambda \ge .95$) &               	&   .671 / .612 / .741  \\ \hline\hline

\multicolumn{4}{c}{Ensemble of (r4, r7, r12) within Fold1}\\
r22   &     majority-voting  $|\mathcal M| > 50\%$           &  {\bf  .669} / .641 / .699 	&   .660 / .663 / .656  \\
r23  &   relax-voting, $|\mathcal M| \ge 30\%$                &   .650 / .525 / {\bf .852} 	&   .674 / .584 / .797  \\  \hline 

\multicolumn{4}{c}{{\bf Ensemble+} of  (r4, r7, r12) from each {\bf Fold1-5}, i.e., $|\mathcal M| = 15$}  \\
r24   &    majority-voting  $|\mathcal M| > 50\%$           &   		&   .658 /  .671 / .645  \\
r25   &     relax-voting, $|\mathcal M| \ge 40\%$          &               &   .673 / .644 / .705  \\
r26   &    relax-voting, $|\mathcal M| \ge 30\%$          &               &    {\bf .679} / .622 /  .747  \\
r27   & \   + postprocess ($w$=10, $\lambda \ge .99$) &               &   .674 / .615 / .747  \\
r28   & \   + postprocess ($w$=10, $\lambda \ge .95$) &               &   .676 / .615 / {\bf .751}  \\ \hline
\end{tabular}}
\caption{SLC: Scores on Dev (internal) of Fold1 and Dev (external) using different classifiers and features.}
\label{tab:SLCDevScores}
\end{table}

\begin{table}[t]
\center
\renewcommand*{\arraystretch}{1.25}
\resizebox{.499\textwidth}{!}{
\setlength\tabcolsep{3.pt}
\begin{tabular}{l|c|c}
\hline
 \multicolumn{2}{c|}{\bf Dev (internal), Fold1}				&   \multicolumn{1}{c}{\bf Dev (external)} \\ 
\multicolumn{1}{c|}{\bf Features}   						& $F1$ \ / \    $P$ \  / \   $R$   &     $F1$  \  /  \    $P$ \ /   \   $R$ \\ \hline
(I) {\it LSTM-CRF} + 	FastTextWordEmb	     				&     .153 / .228 / .115 			&   .122 / .248 / .081 \\		 			
(II)   + Polarity, POS, NER			     					&   .158 / .292	 / .102			&   .101 / .286	 / .061  \\ 
(III)   + Multi-grain (SLC+FLC)		     					&    .148 / .215 / .112 			&   .119 / .200 / .085 \\
(IV)   + BERTSentEmb		     							&   .152 / .264	 / .106			&   .099 / .248	 / .062  \\ 
(V) \  + Multi-task (PFD)			     					&   .144 / .187 / .117 				&   .114 / .179 / .083 \\			 \hline

\multicolumn{3}{c}{Ensemble of  (II and IV) within {\bf Fold1}} \\ 
+ postprocess					 					&   		          			&   .116 / .221	 / .076  \\ \hline\hline

\multicolumn{3}{c}{Ensemble of  (II and IV) within {\bf Fold2}} \\ 
+ postprocess					 					&   		          			&   .129 / .261	 / .085  \\ \hline

\multicolumn{3}{c}{Ensemble of  (II and IV) within {\bf Fold3}} \\ 
+ postprocess					 					&   		       			&   .133 / .220	 / .095  \\ \hline \hline

\multicolumn{3}{c}{{\bf Ensemble+} of  (II and IV) from each {\bf Fold1-3}, i.e., $|\mathcal M| = 6$}  \\ 
+ postprocess					 					&   				            			&   .164 / .182	 / .150  \\ \hline
\end{tabular}}
\caption{FLC: Scores on Dev (internal) of Fold1 and Dev (external) with different models, features and ensembles. PFD: Propaganda Fragment Detection.}
\label{tab:FLCDevScores}
\end{table}

\subsection{Results: Sentence-Level Propaganda}

Table \ref{tab:SLCDevScores} shows the scores on dev (internal and external) for SLC task. 
Observe that the pre-trained embeddings (FastText or BERT) outperform TF-IDF vector representation. 
In row r2, we apply logistic regression classifier with {\it BERTSentEmb} that leads to improved scores over {\it FastTextSentEmb}. 
Subsequently, we augment the sentence vector with additional features that improves F1 on dev (external), however not dev (internal). 
Next, we initialize CNN by {\it FastTextWordEmb} or {\it BERTWordEmb} and augment the last hidden layer (before classification) with {\it BERTSentEmb} and feature vectors, 
leading to gains in F1 for both the dev sets.  Further, we fine-tune BERT and apply different thresholds in relaxing the decision boundary, where $\tau  \ge 0.35$ is found optimal. 

We choose the three different models in the ensemble: Logistic Regression, CNN and BERT on fold1 and 
subsequently an ensemble+ of r3, r6 and r12 from each fold1-5 (i.e., 15 models) to obtain predictions for dev (external).   
We investigate different ensemble schemes (r17-r19), where we observe that the relax-voting improves recall and therefore, the higher F1 (i.e., 0.673). 
In {\it postprocess} step, we check for {\it repetition} propaganda technique by computing cosine similarity between the current sentence and its preceding $w=10$ sentence vectors (i.e., BERTSentEmb) in the document. 
If the cosine-similarity is greater than $\lambda \in \{.99, .95\}$, then the current sentence is labeled as propaganda due to repetition.  
Comparing r19 and r21, we observe a gain in recall, however an overall decrease in F1 applying {\it postprocess}.  

Finally, we use the configuration of r19 on the test set. The ensemble+ of (r4, r7 r12) was analyzed after test submission. 
Table \ref{tab:SLCFLCTestScores} (SLC)  shows that our submission is ranked at $4^{th}$ position.   

\subsection{Results: Fragment-Level Propaganda} 

Table \ref{tab:FLCDevScores} shows the scores on dev (internal and external) for FLC task. 
Observe that the features (i.e., polarity, POS and NER in row II) when introduced in LSTM-CRF improves F1.
We run multi-grained LSTM-CRF without BERTSentEmb (i.e., row III) and with it (i.e., row IV), where 
the latter improves scores on dev (internal), however not on dev (external).  Finally, we perform multi-tasking with another auxiliary task of PFD.    
Given the scores on dev (internal and external) using different configurations (rows I-V), it is difficult to infer the optimal configuration. 
Thus, we choose the two best configurations (II and IV) on dev (internal) set and build an ensemble+ of predictions (discussed in section \ref{sec:FLC}), 
leading to a boost in recall and thus an improved F1 on dev (external).   

Finally, we use the ensemble+ of (II and IV) from each of the folds 1-3, i.e., $|{\mathcal M}|=6$ models to obtain predictions on test. 
Table \ref{tab:SLCFLCTestScores} (FLC) shows that our submission is ranked at $3^{rd}$ position.

\section{Conclusion and Future Work}

Our system (Team: {MIC-CIS}) explores different neural architectures (CNN, BERT and LSTM-CRF) with linguistic, layout and topical features to address 
the tasks of fine-grained propaganda detection. We have demonstrated gains in performance due to the features, ensemble schemes, multi-tasking and multi-granularity architectures.  
Compared to the other participating systems, our submissions are ranked $3^{rd}$ and $4^{th}$ in FLC and SLC tasks, respectively. 

In future, we would like to enrich BERT models with linguistic, layout and topical features during their fine-tuning. 
Further, we would also be interested in understanding and analyzing the neural network learning, i.e.,   
extracting salient fragments (or key-phrases) in the sentence that generate propaganda, similar to \newcite{pankajgupta:2018LISA} in order to promote explainable AI.

\bibliography{emnlp-ijcnlp-2019}
\bibliographystyle{acl_natbib}

\end{document}